%% file: main.tex
\pdfoutput=1
\documentclass[11pt]{article}
\usepackage{acl}
\usepackage{times}
\usepackage{latexsym}

\usepackage{microtype}

\usepackage{subcaption}

\input{settings.tex}

\input{math_commands.tex}

\newcommand\our{\makebox{\textsc{XLM-E}}}

\newcommand\ele{\textsc{ELECTRA}}

\usepackage{pifont}

\title{XLM-E: Cross-lingual Language Model Pre-training via ELECTRA}

\newcommand*\samethanks[1][\value{footnote}]{\footnotemark[#1]}

\author{Zewen Chi$^{\dag\ddag}$\thanks{\ \ Equal contribution. Zewen Chi contributes during internship at Microsoft Research.},~~Shaohan Huang$^\ddag\samethanks[1]$,~~Li Dong$^\ddag$,~~Shuming Ma$^\ddag$,~~Bo Zheng$^\ddag$\textbf{,}
\textbf{Saksham Singhal}$^\ddag$\\~~\textbf{Payal Bajaj}$^\ddag$\textbf{,}~~\textbf{Xia Song}$^\ddag$\textbf{,}~~\textbf{Xian-Ling Mao}$^\dag$\textbf{,}~~\textbf{Heyan Huang}$^\dag$\textbf{,}~~\textbf{Furu Wei}$^\ddag$\\
$^\dag$~Beijing Institute of Technology \\
$^\ddag$~Microsoft Corporation \\
\url{https://github.com/microsoft/unilm} \\}

\date{}

\begin{document}
\maketitle
\begin{abstract}
In this paper, we introduce ELECTRA-style tasks~\cite{electra} to cross-lingual language model pre-training. Specifically, we present two pre-training tasks, namely multilingual replaced token detection, and translation replaced token detection. Besides, we pretrain the model, named as \our{}, on both multilingual and parallel corpora. Our model outperforms the baseline models on various cross-lingual understanding tasks with much less computation cost. Moreover, analysis shows that \our{} tends to obtain better cross-lingual transferability.
\end{abstract}

\section{Introduction}
\label{sec:intro}

\begin{figure}[t]
\centering
\includegraphics[scale=0.5]{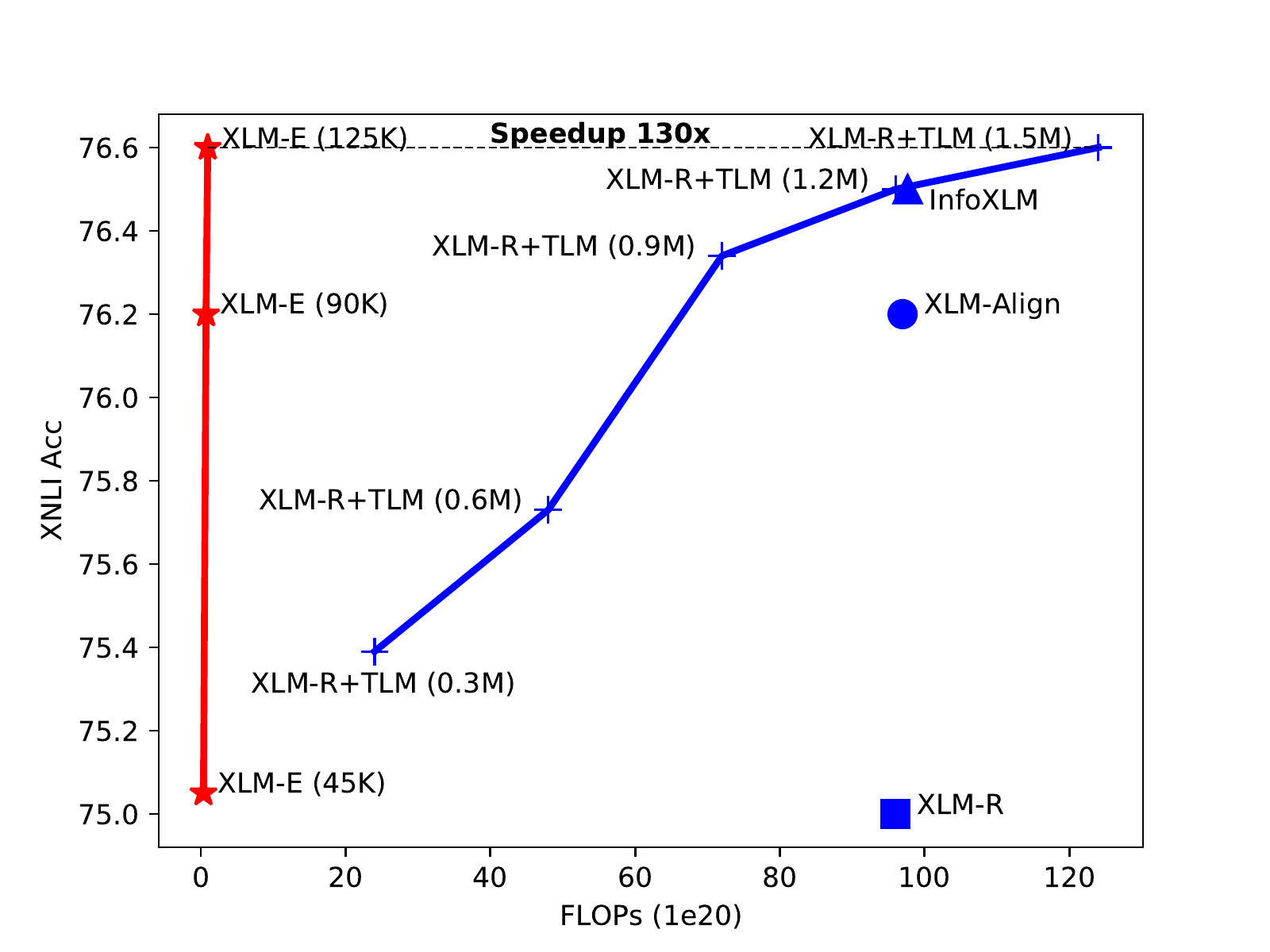}
\caption{The proposed XLM-E pre-training (red line) achieves 130$\times$ speedup compared with an in-house pretrained XLM-R augmented with translation language modeling (XLM-R + TLM; blue line), using the same corpora and code base. The training steps are shown in the brackets. We also present XLM-R~\cite{xlmr}, InfoXLM~\cite{infoxlm}, and XLM-Align~\cite{xlmalign}. The compared models are all in Base size.}
\label{fig:speedup}
\end{figure}

It has become a de facto trend to use a pretrained language model~\cite{bert,unilm,xlnet,unilmv2} for downstream NLP tasks.
These models are typically pretrained with masked language modeling objectives, which learn to generate the masked tokens of an input sentence.
In addition to monolingual representations, the masked language modeling task is effective for learning cross-lingual representations.
By only using multilingual corpora, such pretrained models perform well on zero-shot cross-lingual transfer~\cite{bert,xlmr}, i.e., fine-tuning with English training data while directly applying the model to other target languages.
The cross-lingual transferability can be further improved by introducing external pre-training tasks using parallel corpus, such as translation language modeling~\cite{xlm}, and cross-lingual contrast~\cite{infoxlm}.
However, previous cross-lingual pre-training based on masked language modeling usually requires massive computation resources, rendering such models quite expensive.
As shown in Figure~\ref{fig:speedup}, our proposed XLM-E achieves a huge speedup compared with well-tuned pretrained models.

In this paper, we introduce ELECTRA-style tasks~\cite{electra} to cross-lingual language model pre-training.
Specifically, we present two discriminative pre-training tasks, namely multilingual replaced token detection, and translation replaced token detection. 
Rather than recovering masked tokens, the model learns to distinguish the replaced tokens in the corrupted input sequences.
The two tasks build input sequences by replacing tokens in multilingual sentences, and translation pairs, respectively. 
We also describe the pre-training algorithm of our model, \our{}, which is pretrained with the above two discriminative tasks.
It provides a more compute-efficient and sample-efficient way for cross-lingual language model pre-training.

We conduct extensive experiments on the XTREME cross-lingual understanding benchmark to evaluate and analyze \our{}. Over seven datasets, our model achieves competitive results with the baseline models, while only using 1\% of the computation cost comparing to XLM-R.
In addition to the high computational efficiency, our model also shows the cross-lingual transferability that achieves a reasonably low transfer gap.
We also show that the discriminative pre-training encourages universal representations, making the text representations better aligned across different languages.

Our contributions are summarized as follows:
\begin{itemize}
\item We explore ELECTRA-style tasks for cross-lingual language model pre-training, and pretrain \our{} with both multilingual corpus and parallel data.
\item We demonstrate that \our{} greatly reduces the computation cost of cross-lingual pre-training.
\item We show that discriminative pre-training tends to encourage better cross-lingual transferability.
\end{itemize}

\section{Background: \ele{}}
\label{sec:ele}

\ele{}~\cite{electra} introduces the replaced token detection task for language model pre-training, with the goal of distinguishing real input tokens from corrupted tokens.
That means the text encoders are pretrained as discriminators rather than generators, which is different from the previous pretrained language models, such as BERT~\cite{bert}, that learn to predict the masked tokens.
The ELECTRA pre-training task has shown good performance on various data, such as language~\cite{hao-etal-2021-learning}, and vision~\cite{Fang2022CorruptedIM}.

\ele{} trains two Transformer~\cite{transformer} encoders, serving as generator and discriminator, respectively.
The generator $G$ is typically a small BERT model trained with the masked language modeling (MLM;~\citealt{bert}) task. Consider an input sentence
$\vx = \{x_i\}_{i=1}^{n}$
containing $n$ tokens. MLM first randomly selects a subset $\mathcal{M} \subseteq \left \{ 1, \dots, n \right\}$ as the positions to be masked, and construct the masked sentence $\vx^{\text{masked}}$ by replacing tokens in $\mathcal{M}$ with \texttt{[MASK]}.
Then, the generator predicts the probability distributions of the masked tokens $p_G(x|\vx^{\text{masked}})$. The loss function of the generator $G$ is:
\begin{align}
\Ls_G(\vx; \vtheta_G) = - \sum_{i \in \mathcal{M}} \log p_G(x_i | \vx^{\text{masked}}).
\end{align}
The discriminator $D$ is trained with the replaced token detection task. Specifically, the discriminator takes the corrupted sentences $\vx^{\text{corrupt}}$ as input, which is constructed by replacing the tokens in $\mathcal{M}$ with the tokens sampled from the generator $G$:
\begin{align}
\begin{cases}
x_i^{\text{corrupt}} \sim p_G(x_i|\vx^{\text{masked}}) , & i \in \mathcal{M} \\
x_i^{\text{corrupt}} = x_i, & i \not\in \mathcal{M}
\end{cases}
\end{align}
Then, the discriminator predicts whether $x_i^{\text{corrupt}}$ is original or sampled from the generator. The loss function of the discriminator $D$ is
\begin{align}
\Ls_D(\vx; \vtheta_D) = - \sum_{i = 1}^n \log p_D(z_i|\vx^{\text{corrupt}})
\end{align}
where $z_i$ represents the label of whether $x_i^{\text{corrupt}}$ is the original token or the replaced one.
The final loss function of \ele{} is the combined loss of the generator and discriminator losses, $\Ls_E = \Ls_G + \lambda \Ls_D$.

Compared to generative pre-training, ELECTRA uses more model parameters and training FLOPs per step, because it contains a generator and a discriminator during pre-training.
However, only the discriminator is used for fine-tuning on downstream tasks, so the size of the final checkpoint is similar to BERT-like models in practice.

\section{Methods}
\label{sec:methods}

\begin{figure*}[t]
\centering
\includegraphics[width=1.0\textwidth]{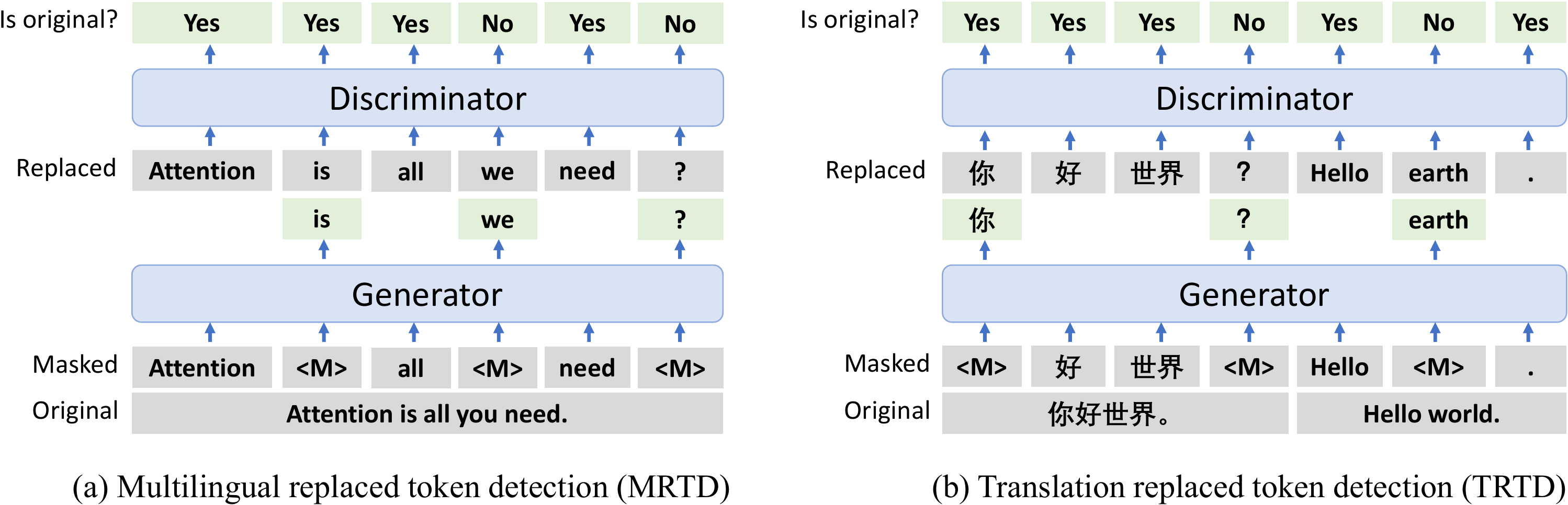}
\caption{Overview of two pre-training tasks of \our{}, i.e., multilingual replaced token detection, and translation replaced token detection. The generator predicts the masked tokens given a masked sentence or a masked translation pair, and the discriminator distinguishes whether the tokens are replaced by the generator.}
\label{fig:rtd}
\end{figure*}

Figure~\ref{fig:rtd} shows an overview of the two discriminative tasks used for pre-training \our{}.
Similar to \ele{} described in Section~\ref{sec:ele}, \our{} has two Transformer components, i.e., generator and discriminator. The generator predicts the masked tokens given the masked sentence or translation pair, and the discriminator distinguishes whether the tokens are replaced by the generator.

\subsection{Pre-training Tasks}

The pre-training tasks of \our{} are multilingual replaced token detection (MRTD), and translation replaced token detection (TRTD).

\paragraph{Multilingual Replaced Token Detection}

The multilingual replaced token detection task requires the model to distinguish real input tokens from corrupted multilingual sentences.
Both the generator and the discriminator are shared across languages.
The vocabulary is also shared for different languages.
The task is the same as in monolingual ELECTRA pre-training (Section~\ref{sec:ele}). The only difference is that the input texts can be in various languages.

We use uniform masking to produce the corrupted positions.
We also tried span masking~\cite{spanbert,unilmv2} in our preliminary experiments. The results indicate that span masking significantly weakens the generator's prediction accuracy, which in turn harms pre-training.

\paragraph{Translation Replaced Token Detection}

Parallel corpora are easily accessible and proved to be effective for learning cross-lingual language models~\cite{xlm,infoxlm}, while it is under-studied how to improve discriminative pre-training with parallel corpora. We introduce the translation replaced token detection task that aims to distinguish real input tokens from translation pairs. Given an input translation pair, the generator predicts the masked tokens in both languages. Consider an input translation pair $(\ve, \vf)$. We construct the input sequence by concatenating the translation pair as a single sentence.
The loss function of the generator $G$ is:
\begin{align}
\nonumber
\Ls_G(\ve, \vf; \vtheta_G) =& - \sum_{i \in \mathcal{M}_e} \log p_G(e_i | \left[ \ve ; \vf \right]^{\text{masked}}) \\
\nonumber
& - \sum_{i \in \mathcal{M}_f} \log p_G(f_i | \left[ \ve ; \vf \right]^{\text{masked}})
\end{align}
where $\left[ ; \right]$ is the operator of concatenation, and $\mathcal{M}_e, \mathcal{M}_f$ stand for the randomly selected masked positions for $\ve$ and $\vf$, respectively. This loss function is identical to the translation language modeling loss (TLM;~\citealt{xlm}).
The discriminator $D$ learns to distinguish real input tokens from the corrupted translation pair. The corrupted translation pair $(\ve^{\text{corrupt}}, \vf^{\text{corrupt}})$ is constructed by replacing tokens with the tokens sampled from $G$ with the concatenated translation pair as input. Formally, $\ve^{\text{corrupt}}$ is constructed by
\begin{align}
\begin{cases}
e_i^{\text{corrupt}} \sim p_G(e_i | \left[ \ve ; \vf \right]^{\text{masked}}) , & i \in \mathcal{M}_e \\
e_i^{\text{corrupt}} = e_i, & i \not\in \mathcal{M}_e
\end{cases}
\end{align}
The same operation is also used to construct $\vf^{\text{corrupt}}$. Then, the loss function of the discriminator $D$ can be written as
\begin{align}
\Ls_D(\ve, \vf; \vtheta_D) = - \sum_{i = 1}^{n_e+n_f} \log p_D(r_i| \left[ \ve ; \vf \right]^{\text{corrupt}})
\end{align}
where $r_i$ represents the label of whether the $i$-th input token is the original one or the replaced one. The final loss function of the translation replaced token detection task is $\Ls_G + \lambda \Ls_D$.

\subsection{Pre-training \our{}}

The \our{} model is jointly pretrained with the masked language modeling, translation language modeling, multilingual replaced token detection and the translation replaced token detection tasks. The overall training objective is to minimize
\begin{align}
\nonumber
\Ls~=~&\Ls_\text{MLM}(\vx;\theta_G) + \Ls_\text{TLM}(\ve, \vf;\theta_G) \\ \nonumber
&+ \lambda \Ls_\text{MRTD}(\vx;\theta_D) + \lambda \Ls_\text{TRTD}(\ve, \vf; \theta_D)
\end{align}
over large scale multilingual corpus $\mathcal{X} = \left\{ \vx \right\}$ and parallel corpus $\mathcal{P} = \left\{ (\ve, \vf) \right\}$.
We jointly pretrain the generator and the discriminator from scratch.
Following \citet{electra}, we make the generator smaller to improve the pre-training efficiency. 

\subsection{Gated Relative Position Bias}
\label{sec:grep}

We propose to use gated relative position bias in the self-attention mechanism.
Given input tokens $\{x_i\}_{i=1}^{|x|}$, let $\{\mathbf{h}_i\}_{i=1}^{|x|}$ denote their hidden states in Transformer.
The self-attention outputs $\{\tilde{\mathbf{h}}_i\}_{i=1}^{|x|}$ are computed via:
\begin{align}
\mathbf{q}_i,\mathbf{k}_i,\mathbf{v}_i &= \mathbf{h}_i \mathbf{W}^Q, \mathbf{h}_i \mathbf{W}^K, \mathbf{h}_i \mathbf{W}^V \\
a_{ij} &\propto \exp\{ \frac{ \mathbf{q}_i \cdot \mathbf{k}_j }{\sqrt{d_{k}}} + r_{i-j} \} \\
\tilde{\mathbf{h}}_i &= \sum_{j=1}^{|x|}{ a_{ij} \mathbf{v}_i }
\end{align}
where $r_{i-j}$ represents gated relative position bias, each $\mathbf{h}_i$ is linearly projected to a triple of query, key and value using parameter matrices $\mathbf{W}^Q , \mathbf{W}^K , \mathbf{W}^V \in \mathbb{R}^{d_h \times d_k}$, respectively.

Inspired by the gating mechanism of Gated Recurrent Unit (GRU; \citealt{gru}), we compute gated relative position bias $r_{i-j}$ via:
\begin{align}
g^\text{(update)} , g^\text{(reset)} &= \sigmoid ( \mathbf{q}_i \cdot \mathbf{u} ), \sigmoid ( \mathbf{q}_i \cdot \mathbf{v} ) \nonumber \\
\tilde{r}_{i-j} &= w g^\text{(reset)} d_{i-j} \nonumber \\
r_{i-j} = d_{i-j} + g^\text{(update)} &d_{i-j} + (1 - g^\text{(update)}) \tilde{r}_{i-j} \nonumber
\end{align}
where $d_{i-j}$ is learnable relative position bias, the vectors $\mathbf{u},\mathbf{v} \in \mathbb{R}^{d_k}$ are parameters, $\sigmoid$ is a sigmoid function, and $w$ is a learnable value.

Compared with relative position bias~\cite{parikh-etal-2016-decomposable,t5,unilmv2}, the proposed gates take the content into consideration, which adaptively adjusts the relative position bias by conditioning on input tokens.
Intuitively, the same distance between two tokens tends to play different roles in different languages.

\subsection{Initialization of Transformer Parameters}
\label{sec:init}

Properly initializing Transformer parameters is critical to stabilize large-scale training.
First, all the parameters are randomly initialized by uniformly sampling from a small range, such as $[-0.02, 0.02]$.
Second, for the $l$-th Transformer block\footnote{Each block contains a self-attention layer and a feed-forward network layer.}, we rescale the attention output weight and the feed-forward network output matrix by $1/\sqrt{2l}$. Notice that the Transformer block after the embedding layer is regarded the first one.

\section{Experiments}
\label{sec:exp}

\subsection{Setup}

\paragraph{Data}

We use the CC-100~\cite{xlmr} dataset for the replaced token detection task. CC-100 contains texts in 100 languages collected from the CommonCrawl dump. We use parallel corpora for the translation replaced token detection task, including translation pairs in 100 languages collected from MultiUN~\cite{multiun}, IIT Bombay~\cite{iit}, OPUS~\cite{opus}, WikiMatrix~\cite{wikimatrix}, and CCAligned~\cite{el2019ccaligned}.

Following XLM~\cite{xlm}, we sample multilingual sentences to balance the language distribution. Formally, consider the pre-training corpora in $N$ languages with $m_j$ examples for the $j$-th language.
The probability of using an example in the $j$-th language is
\begin{align}
p_j = \frac{m_j^\alpha}{\sum_{k=1}^N m_k^\alpha}
\end{align}
The exponent $\alpha$ controls the distribution such that a lower $\alpha$ increases the probability of sampling examples from a low-resource language. In this paper, we set $\alpha=0.7$.

\paragraph{Model}

We use a Base-size $12$-layer Transformer~\cite{transformer} as the discriminator, with hidden size of $768$, and FFN hidden size of $3,072$.
The generator is a $4$-layer Transformer using the same hidden size as the discriminator~\cite{cocolm}.
See Appendix~\ref{appendix:params_model} for more details of model hyperparameters.

\paragraph{Training}

We jointly pretrain the generator and the discriminator of \our{} from scratch, using the Adam~\cite{adam} optimizer for 125K training steps.
We use dynamic batching of approximately 1M tokens for each pre-training task.
We set $\lambda$, the weight for the discriminator objective to 50.
The whole pre-training procedure takes about 1.7 days on 64 Nvidia A100 GPU cards. See Appendix~\ref{appendix:params_pretrain} for more details of pre-training hyperparameters.

\begin{table*}[t]
\centering
\scalebox{0.81}{
\renewcommand\tabcolsep{5.0pt}
\begin{tabular}{lcccccccccccccccccc}
\toprule
\multirow{2}{*}{\bf Model} & \multicolumn{2}{c}{\bf Structured Prediction} & \multicolumn{3}{c}{\bf Question Answering} & \multicolumn{2}{c}{\bf Classification} & \multirow{2}{*}{\bf Avg} \\
& POS & NER & XQuAD & MLQA & TyDiQA & XNLI & PAWS-X & \\ \midrule
Metrics & F1 & F1 & F1 / EM & F1 / EM & F1 / EM & Acc. & Acc. & \\ 
\midrule
\multicolumn{9}{l}{~~\textit{Pre-training on multilingual corpus}} \\
\textsc{mBert}~\cite{xtreme} & 70.3 & 62.2 & 64.5 / 49.4 & 61.4 / 44.2 & 59.7 / 43.9 & 65.4 & 81.9 & 63.1 \\
\textsc{mT5}~\cite{mt5} & - & 55.7 & 67.0 / 49.0 & 64.6 / 45.0 & 57.2 / 41.2 & \bf 75.4 &  86.4 & - \\
\textsc{XLM-R} & \bf 75.6 & 61.8 & 71.9 / 56.4 & 65.1 / 47.2 & 55.4 / 38.3 & 75.0 & 84.9 & 66.4 \\
\our{} (w/o TRTD) & 74.2 & \bf 62.7 & \bf 74.3 / 58.2 & \bf 67.8 / 49.7 & \bf 57.8 / 40.6 & 75.1 & \bf 87.1 & \bf 67.6 \\ \midrule
\multicolumn{9}{l}{~~\textit{Pre-training on both multilingual corpus and parallel corpus}} \\
\textsc{XLM}~\cite{xtreme}  & 70.1 & 61.2 & 59.8 / 44.3 & 48.5 / 32.6 & 43.6 / 29.1 & 69.1 & 80.9 & 58.6 \\
\textsc{InfoXLM}~\cite{infoxlm} & - & - & -~~~/~~~- & 68.1 / 49.6 & -~~~/~~~- & 76.5 & - & - \\
\textsc{XLM-Align}~\cite{xlmalign} & \bf 76.0 & \bf 63.7 & 74.7 / 59.0 & 68.1 / \textbf{49.8} &  62.1 / 44.8 &  76.2 & 86.8 & 68.9 \\
\our{} &  75.6 & 63.5 & \bf 76.2 / 60.2 & \bf 68.3 / 49.8 & \bf 62.4 / 45.7 & \bf 76.6 & \bf 88.3 & \bf 69.3 \\
\bottomrule
\end{tabular}}
\caption{Evaluation results on XTREME cross-lingual understanding tasks. We consider the cross-lingual transfer setting, where models are only fine-tuned on the English training data but evaluated on all target languages.
The compared models are all in Base size.
Results of \our{} and XLM-R are averaged over five runs.}
\label{table:overview}
\end{table*}

\subsection{Cross-lingual Understanding}

We evaluate \our{} on the XTREME~\cite{xtreme} benchmark, which is a multilingual multi-task benchmark for evaluating cross-lingual understanding. The XTREME benchmark contains seven cross-lingual understanding tasks, namely part-of-speech tagging on the Universal Dependencies v2.5~\cite{udpos}, NER named entity recognition on the Wikiann~\cite{panx,rahimi2019} dataset, cross-lingual natural language inference on XNLI~\cite{xnli}, cross-lingual paraphrase adversaries from word scrambling (PAWS-X;~\citealt{pawsx}), and cross-lingual question answering on MLQA~\cite{mlqa}, XQuAD~\cite{xquad}, and TyDiQA-GoldP~\cite{tydiqa}.

\paragraph{Baselines}

We compare our \our{} model with the cross-lingual language models pretrained with multilingual text, i.e., Multilingual BERT (\textsc{mBert};~\citealt{bert}), \textsc{mT5}~\cite{mt5}, and XLM-R~\cite{xlmr}, or pretrained with both multilingual text and parallel corpora, i.e., XLM~\cite{xlm}, \textsc{InfoXLM}~\cite{infoxlm}, and \textsc{XLM-Align}~\cite{xlmalign}. 
The compared models are all in Base size. In what follows, models are considered as in Base size by default.

\paragraph{Results}

We use the cross-lingual transfer setting for the evaluation on XTREME~\cite{xtreme}, where the models are first fine-tuned with the English training data and then evaluated on the target languages.
In Table~\ref{table:overview}, we report the accuracy, F1, or Exact-Match (EM) scores on the XTREME cross-lingual understanding tasks. The results are averaged over all target languages and five runs with different random seeds.
We divide the pretrained models into two categories, i.e., the models pretrained on multilingual corpora, and the models pretrained on both multilingual corpora and parallel corpora. For the first setting, we pretrain \our{} with only the multilingual replaced token detection task.
From the results, it can be observed that \our{} outperforms previous models on both settings, achieving the averaged scores of 67.6 and 69.3, respectively. Compared to \textsc{XLM-R}, \our{} (w/o TRTD) produces an absolute 1.2 improvement on average over the seven tasks. For the second setting, compared to \textsc{XLM-Align}, \our{} produces an absolute 0.4 improvement on average. 
\our{} performs better on the question answering tasks and sentence classification tasks while preserving reasonable high F1 scores on structured prediction tasks. 
Despite the effectiveness of \our{}, our model requires substantially lower computation cost than \textsc{XLM-R} and \textsc{XLM-Align}. A detailed efficiency analysis in presented in Section~\ref{sec:eff}.

\begin{table}[t]
\centering
\small
\scalebox{0.95}{
\renewcommand\tabcolsep{3.5pt}
\begin{tabular}{lcc}
\toprule
\bf Model & \bf XNLI & \bf MLQA \\
\midrule
XLM (reimplementation) & 73.4 & 66.2 / 47.8  \\
~~~$-$TLM & 70.6 & 64.0 / 46.0 \\ \midrule
\our{} & \bf 76.6 & \bf 68.3 / 49.8 \\
~~~$-$TRTD & 75.1 & 67.8 / 49.7  \\ 
~~~$-$TRTD$-$Gated relative position bias &  75.2 & 67.4 / 49.2 \\
\bottomrule
\end{tabular}
}
\caption{Ablation studies of \our{}. We studies the effects of the main components of \our{}, and compare the models with XLM under the same pre-training setup, including training steps, learning rate, etc.}
\label{table:ablation}
\end{table}

\subsection{Ablation Studies}

For a deeper insight to \our{}, we conduct ablation experiments where we first remove the TRTD task and then remove the gated relative position bias. Besides, we reimplement XLM that is pretrained with the same pre-training setup with \our{}, i.e., using the same training steps, learning rate, etc. Table~\ref{table:ablation} shows the ablation results on XNLI and MLQA. Removing TRTD weakens the performance of \our{} on both downstream tasks. On this basis, the results on MLQA further decline when removing the gated relative position bias. This demonstrates that \our{} benefits from both TRTD and the gated relative position bias during pre-training.
Besides, \our{} substantially outperform XLM on both tasks. Notice that when removing the two components from \our{}, our model only requires a multilingual corpus, but still achieves better performance than XLM, which uses an additional parallel corpus.

\begin{table}[t]
\centering
\small
\begin{tabular}{lcccc}
\toprule
\bf Model & \bf Size & \bf Params & \bf XNLI & \bf MLQA \\
\midrule
\our{} & Base & 279M & 76.6 & 68.3 / 49.8 \\
\our{} & Large & 840M & 81.3 & 72.7 / 54.2 \\
\our{} & XL & 2.2B & \bf 83.7 & \bf 76.2 / 57.9 \\ \midrule
XLM-R & XL  & 3.5B & 82.3 & 73.4 / 55.3 \\
\textsc{mT5} & XL  & 3.7B & 82.9 & 73.5 / 54.5 \\
\bottomrule
\end{tabular}
\caption{Results of scaling-up the model size.}
\label{table:scaleup}
\end{table}

\subsection{Scaling-up Results}
\label{sec:scaleup}

Scaling-up model size has shown to improve performance on cross-lingual downstream tasks~\cite{mt5,xlmr-scaleup}.
We study the scalability of \our{} by pre-training \our{} models using larger model sizes. 
We consider two larger model sizes in our experiments, namely Large and XL. Detailed model hyperparameters can be found in Appendix~\ref{appendix:params_model}.
As present in Table~\ref{table:scaleup}, XLM-E$_\text{XL}$ achieves the best performance while using significantly fewer parameters than its counterparts. Besides, scaling-up the \our{} model size consistently improves the results, demonstrating the effectiveness of \our{} for large-scale pre-training.

\subsection{Training Efficiency}
\label{sec:eff}

\begin{table}[t]
\centering
\small
\renewcommand\tabcolsep{3.5pt}
\begin{tabular}{lccc}
\toprule
\bf Model & \bf XTREME & \bf Params & \bf FLOPs \\
\midrule
\textsc{mBERT} & 63.1 & 167M & 6.4e19 \\
XLM-R & 66.4 & 279M & 9.6e21 \\
\textsc{InfoXLM}* & - & 279M & 9.6e21 + 1.7e20 \\
\textsc{XLM-Align}* & 68.9 & 279M & 9.6e21 + 9.6e19 \\
\our{} & 69.3 & 279M & 9.5e19 \\
~~~$-$TRTD & 67.6 & 279M & 6.3e19 \\
\bottomrule
\end{tabular}
\caption{Comparison of the pre-training costs.
The models with `*' are continue-trained from XLM-R rather than pre-training from scratch.}
\label{table:eff}
\end{table}

We present a comparison of the pre-training resources, to explore whether \our{} provides a more compute-efficient and sample-efficient way for pre-training cross-lingual language models.
Table~\ref{table:eff} compares the XTREME average score, the number of parameters, and the pre-training computation cost.
Notice that \textsc{InfoXLM} and \textsc{XLM-Align} are continue-trained from XLM-R, so the total training FLOPs are accumulated over XLM-R.

Table~\ref{table:eff} shows that \our{} substantially reduces the computation cost for cross-lingual language model pre-training.
Compared to XLM-R and \textsc{XLM-Align} that use at least 9.6e21 training FLOPs, \our{} only uses 9.5e19 training FLOPs in total while even achieving better XTREME performance than the two baseline models.
For the setting of pre-training with only multilingual corpora, \our{} (w/o TRTD) also outperforms XLM-R using 6.3e19 FLOPs in total.
This demonstrates the compute-effectiveness of \our{}, i.e., \our{} as a stronger cross-lingual language model requires substantially less computation resource.

\subsection{Cross-lingual Alignment}

To explore whether discriminative pre-training improves the resulting cross-lingual representations, we evaluate our model on the sentence-level and word-level alignment tasks, i.e., cross-lingual sentence retrieval and word alignment.

\begin{table}
\centering
\small
\scalebox{0.94}{
\renewcommand\tabcolsep{5.0pt}
\begin{tabular}{lccccc}
\toprule
\multirow{2}{*}{\bf Model} & \multicolumn{2}{c}{\bf Tatoeba-14} & \multicolumn{2}{c}{\bf Tatoeba-36} \\
& en $\rightarrow$ xx & xx $\rightarrow$ en & en $\rightarrow$ xx & xx $\rightarrow$ en \\ \midrule
\textsc{XLM-R} & 59.5 & 57.6 & 55.5 & 53.4 \\
\textsc{InfoXLM} & \bf 80.6 & \bf 77.8 & \bf 68.6 & \bf 67.3 \\
\our{} & 74.4 & 72.3 & 65.0 & 62.3 \\
~~~$-$TRTD & 55.8 & 55.1 & 46.4 & 44.6 \\
\bottomrule
\end{tabular}}
\caption{Average accuracy@1 scores for Tatoeba cross-lingual sentence retrieval. The models are evaluated under two settings with 14 and 36 of the parallel corpora for evaluation, respectively.
}
\label{table:tat}
\end{table}

We use the Tatoeba~\cite{tatoeba} dataset for the cross-lingual sentence retrieval task, the goal of which is to find translation pairs from the corpora in different languages.
Tatoeba consists of English-centric parallel corpora covering 122 languages. Following \citet{infoxlm} and \citet{xtreme}, we consider two settings where we use 14 and 36 of the parallel corpora for evaluation, respectively.
The sentence representations are obtained by average pooling over hidden vectors from a middle layer. Specifically, we use layer-7 for XLM-R and layer-9 for \our{}. Then, the translation pairs are induced by the nearest neighbor search using the cosine similarity.
Table~\ref{table:tat} shows the average accuracy@1 scores under the two settings of Tatoeba for both the xx $\rightarrow$ en and en $\rightarrow$ xx directions. \our{} achieves 74.4 and 72.3 accuracy scores for Tatoeba-14, and 65.0 and 62.3 accuracy scores for Tatoeba-36, providing notable improvement over \textsc{XLM-R}. \our{} performs slightly worse than \textsc{InfoXLM}. We believe the cross-lingual contrast~\cite{infoxlm} task explicitly learns the sentence representations, which makes \textsc{InfoXLM} more effective for the cross-lingual sentence retrieval task.

\begin{table}
\centering
\small
\scalebox{0.98}{
\renewcommand\tabcolsep{5.0pt}
\begin{tabular}{lcccccc}
\toprule
\multirow{2}{*}{\bf Model} & \multicolumn{4}{c}{\bf Alignment Error Rate $\downarrow$} & \multirow{2}{*}{\bf Avg} \\
 & en-de & en-fr & en-hi & en-ro &  \\ \midrule
fast\_align & 32.14 & 19.46 & 59.90 & - & - \\
\textsc{XLM-R} & {17.74} & 7.54 & {37.79} & 27.49 & {22.64} \\
\textsc{XLM-Align} & 16.63 & 6.61 & 33.98 & 26.97 & 21.05 \\
\our{} & \bf 16.49 & \bf 6.19 & \bf 30.20 & \bf 24.41 & \bf 19.32 \\
~~~$-$TRTD & 17.87 & 6.29 & 35.02 & 30.22 & 22.35 \\
\bottomrule
\end{tabular}}
\caption{Alignment error rate scores (lower is better) for the word alignment task on four language pairs. Results of the baseline models are from \citet{xlmalign}. We use the optimal transport method to obtain the resulting word alignments, where the sentence representations are from the $9$-th layer of \our{}.
}
\label{table:wa}
\end{table}

For the word-level alignment, we use the word alignment datasets from EuroParl\footnote{\url{www-i6.informatik.rwth-aachen.de/goldAlignment/}}, WPT2003\footnote{\url{web.eecs.umich.edu/~mihalcea/wpt/}}, and WPT2005\footnote{\url{web.eecs.umich.edu/~mihalcea/wpt05/}}, containing 1,244 translation pairs annotated with golden alignments. The predicted alignments are evaluated by alignment error rate~(AER; \citealt{och2003systematic}):
\begin{align}
  \text{AER}=1-\frac{|A \cap S|+|A \cap P|}{|A|+|S|}
\end{align}
where $A, S,$ and $P$ stand for the predicted alignments, the annotated sure alignments, and the annotated possible alignments, respectively.
In Table~\ref{table:wa} we compare \our{} with baseline models, i.e., fast\_align~\cite{fastalign}, XLM-R, and \textsc{XLM-Align}. The resulting word alignments are obtained by the optimal transport method~\cite{xlmalign}, where the sentence representations are from the $9$-th layer of \our{}. Over the four language pairs, \our{} achieves lower AER scores than the baseline models, reducing the average AER from $21.05$ to 19.32. It is worth mentioning that our model requires substantial lower computation costs than the other cross-lingual pretrained language models to achieve such low AER scores. 
See the detailed training efficiency analysis in Section~\ref{sec:eff}.
It is worth mentioning that \our{} shows notable improvements over \our{} (w/o TRTD) on both tasks, demonstrating that the translation replaced token detection task is effective for cross-lingual alignment.

\subsection{Universal Layer Across Languages}

We evaluate the word-level and sentence-level representations over different layers to explore whether the \our{} tasks encourage universal representations.

\begin{figure}
\centering
\includegraphics[width=0.42\textwidth]{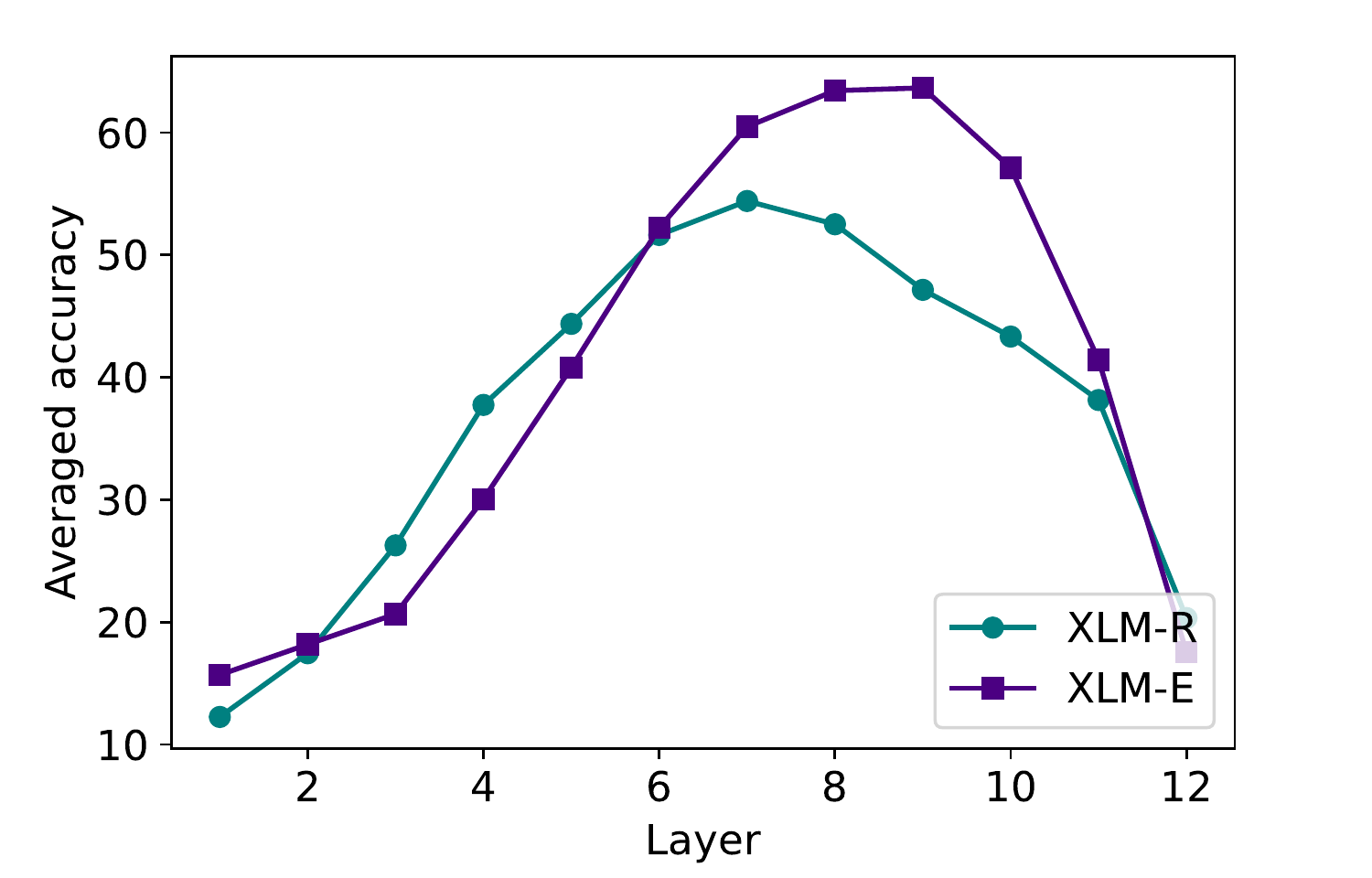}
\caption{Evaluation results on Tatoeba cross-lingual sentence retrieval over different layers. For each layer, the accuracy score is averaged over all the 36 language pairs in both the xx $\rightarrow$ en and en $\rightarrow$ xx directions. }
\label{fig:tat}
\end{figure}

As shown in Figure~\ref{fig:tat}, we illustrate the accuracy@1 scores of \our{} and XLM-R on Tatoeba cross-lingual sentence retrieval, using sentence representations from different layers. For each layer, the final accuracy score is averaged over all the 36 language pairs in both the xx $\rightarrow$ en and en $\rightarrow$ xx directions. 
From the figure, it can be observed that \our{} achieves notably higher averaged accuracy scores than XLM-R for the top layers. The results of \our{} also show a parabolic trend across layers, i.e., the accuracy continuously increases before a specific layer and then continuously drops.
This trend is also found in other cross-lingual language models such as XLM-R and XLM-Align~\cite{simalign,xlmalign}. Different from XLM-R that achieves the highest accuracy of 54.42 at layer-7, \our{} pushes it to layer-9, achieving an accuracy of 63.66. At layer-10, XLM-R only obtains an accuracy of 43.34 while \our{} holds the accuracy score as high as 57.14.

\begin{figure}
\centering
\includegraphics[width=0.42\textwidth]{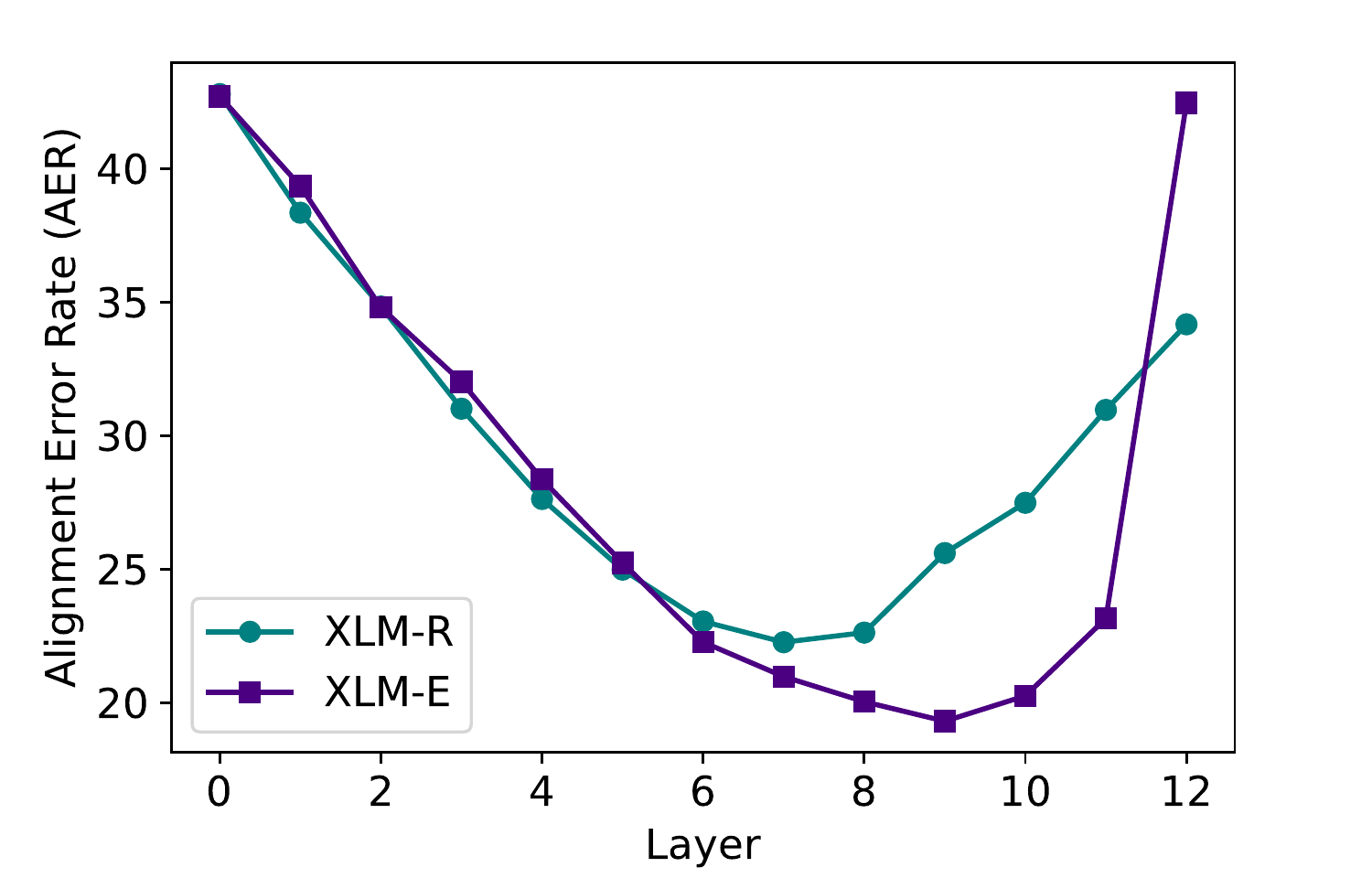}
\caption{Evaluation results of cross-lingual word alignment over different layers. Layer-0 stands for the embedding layer.}
\label{fig:wa}
\end{figure}

Figure~\ref{fig:wa} shows the averaged alignment error rate (AER) scores of \our{} and XLM-R on the word alignment task. We use the hidden vectors from different layers to perform word alignment, where layer-0 stands for the embedding layer. The final AER scores are averaged over the four test sets in different languages. Figure~\ref{fig:wa} shows a similar trend to that in Figure~\ref{fig:tat}, where \our{} not only provides substantial performance improvements over XLM-R, but also pushes the best-performance layer to a higher layer, i.e., the model obtains the best performance at layer-9 rather than a lower layer such as layer-7. 

On both tasks, \our{} shows good performance for the top layers, even though both \our{} and XLM-R use the Transformer~\cite{transformer} architecture. 
Compared to the masked language modeling task that encourages the top layers to be language-specific, discriminative pre-training makes \our{} producing better-aligned text representations at the top layers.
It indicates that the cross-lingual discriminative pre-training encourages universal representations inside the model.

\begin{table}[t]
\centering
\small
\scalebox{0.90}{
\renewcommand\tabcolsep{3.0pt}
\begin{tabular}{lccccc}
\toprule
\bf Model & \textbf{XQuAD} & \textbf{MLQA} & \textbf{TyDiQA} & \bf XNLI & \bf PAWS-X \\ \midrule
\textsc{mBERT} & 25.0 & 27.5 & 22.2 & 16.5 & 14.1 \\
XLM-R & 15.9 & 20.3 & 15.2 & 10.4 & 11.4 \\
\textsc{InfoXLM} & - & 18.8 & - & \bf 10.3 & - \\
\textsc{XLM-Align} & \bf 14.6 & 18.7 & \bf 10.6 & 11.2 & 9.7\\
\our{} & 14.9 & 19.2 & 13.1 & 11.2 & \bf 8.8 \\
~~~$-$TRTD & 16.3 & \bf 18.6 & 16.3 & 11.5 & 9.6 \\
\bottomrule
\end{tabular}}
\caption{The cross-lingual transfer gap scores on the XTREME tasks. A lower transfer gap score indicates better cross-lingual transferability. We use the EM scores to compute the gap scores for the QA tasks.}
\label{table:gap}
\end{table}

\subsection{Cross-lingual Transfer Gap}
We analyze the cross-lingual transfer gap~\cite{xtreme} of the pretrained cross-lingual language models. The transfer gap score is the difference between performance on the English test set and the average performance on the test set in other languages. This score suggests how much end task knowledge has not been transferred to other languages after fine-tuning. A lower gap score indicates better cross-lingual transferability. Table~\ref{table:gap} compares the cross-lingual transfer gap scores on five of the XTREME tasks.
We notice that \our{} obtains the lowest gap score only on PAWS-X. Nonetheless, it still achieves reasonably low gap scores on the other tasks with such low computation cost, demonstrating the cross-lingual transferability of \our{}. 
We believe that it is more difficult to achieve the same low gap scores when the model obtains better performance.

\section{Related Work}

Learning self-supervised tasks on large-scale multilingual texts has proven to be effective for pre-training cross-lingual language models. 
Masked language modeling (MLM;~\citealt{bert}) is typically used to learn cross-lingual encoders such as multilingual BERT (mBERT;~\citealt{bert}) and XLM-R~\cite{xlmr}.
The cross-lingual language models can be further improved by introducing external pre-training tasks using parallel corpora.
XLM~\cite{xlm} introduces the translation language modeling (TLM) task that predicts masked tokens from concatenated translation pairs. ALM~\cite{alm} utilizes translation pairs to construct code-switched sequences as input. 
InfoXLM~\cite{infoxlm} considers an input translation pair as cross-lingual views of the same meaning, and proposes a cross-lingual contrastive learning task.
Several pre-training tasks utilize the token-level alignments in parallel data to improve cross-lingual language models~\cite{Cao2020Multilingual,zhao2020inducing,hu2020explicit,xlmalign}.

In addition, parallel data are also employed for cross-lingual sequence-to-sequence pre-training.
XNLG~\cite{xnlg} presents cross-lingual masked language modeling and cross-lingual auto-encoding for cross-lingual natural language generation, and achieves the cross-lingual transfer for NLG tasks. VECO~\cite{veco} utilizes cross-attention MLM to pretrain a variable cross-lingual language model for both NLU and NLG. mT6~\cite{mt6} improves mT5~\cite{mt5} by learning the translation span corruption task on parallel data. $\Delta$LM~\cite{deltalm} proposes to align pretrained multilingual encoders to improve cross-lingual sequence-to-sequence pre-training.

\section{Conclusion}

We introduce \our{}, a cross-lingual language model pretrained by \ele{}-style tasks.
Specifically, we present two pre-training tasks, i.e., multilingual replaced token detection, and translation replaced token detection.
\our{} outperforms baseline models on cross-lingual understanding tasks although using much less computation cost.
In addition to improved performance and computational efficiency, we also show that \our{} obtains the cross-lingual transferability with a reasonably low transfer gap.

\section{Ethical Considerations}
Our work introduces ELECTRA-style tasks for cross-lingual language model pre-training, which requires much less computation cost than previous models and substantially reduces the energy cost.

\bibliographystyle{acl_natbib}
\bibliography{xlme}

\newpage
\appendix

\section*{Appendix}

\section{Model Hyperparameters}
\label{appendix:params_model}

Table~\ref{table:g-hparam} and Table~\ref{table:d-hparam} shows the model hyperparameters of \our{} in the sizes of Base, Large, and XL. 
For the Base-size model, we use the same vocabulary with XLM-R~\cite{xlmr} that consists of 250K subwords tokenized by SentencePiece~\cite{sentencepiece}. For the models in Large size and XL size, we use VoCap~\cite{vocap} to allocate a 500K vocabulary for models in Large size and XL size.

\begin{table}[ht]
\centering
\small
\renewcommand\tabcolsep{2.8pt}
\begin{tabular}{lrrr}
\toprule
Hyperparameters & Base & Large & XL \\ \midrule
Layers & 4 & 6 & 8 \\
Hidden size & 768 & 1,024 & 1,536 \\
FFN inner hidden size & 3,072 & 4,096 & 6,144 \\
Attention heads & 12 & 16 & 24 \\
\bottomrule
\end{tabular}
\caption{Model hyperparameters of \our{} generators in different sizes.}
\label{table:g-hparam}
\end{table}

\begin{table}[ht]
\centering
\small
\renewcommand\tabcolsep{3.5pt}
\begin{tabular}{lrrr}
\toprule
Hyperparameters & Base & Large & XL \\ \midrule
Layers & 12 & 24 & 48 \\
Hidden size & 768 & 1,024 & 1,536 \\
FFN inner hidden size & 3,072 & 4,096 & 6,144 \\
Attention heads & 12 & 16 & 24 \\
\bottomrule
\end{tabular}
\caption{Model hyperparameters of \our{} discriminators in different sizes.}
\label{table:d-hparam}
\end{table}

\section{Hyperparameters for Pre-Training}
\label{appendix:params_pretrain}
As shown in Table~\ref{table:pt-hparam}, we present the hyperparameters for pre-training \our{}.
We use the batch size of 1M tokens for each pre-training task. In multilingual replaced token detection, a batch is constructed by 2,048 length-512 input sequences, while the input length is dynamically set as the length of the original translation pairs in translation replaced token detection.

\begin{table}[ht]
\centering
\small
\renewcommand\tabcolsep{3.5pt}
\begin{tabular}{lr}
\toprule
Hyperparameters & Value \\ \midrule
Training steps & 125K \\
Batch tokens per task & 1M \\
Adam $\epsilon$ & 1e-6 \\
Adam $\beta$ & (0.9, 0.98) \\
Learning rate & 5e-4 \\
Learning rate schedule & Linear \\
Warmup steps & 10,000 \\
Gradient clipping & 2.0 \\
Weight decay & 0.01 \\
\bottomrule
\end{tabular}
\caption{Hyperparameters used for pre-training \our{}.
}
\label{table:pt-hparam}
\end{table}

\section{Hyperparameters for Fine-Tuning}

In Table~\ref{table:hparam}, we report the hyperparameters for fine-tuning \our{} on the XTREME end tasks.

\begin{table*}[t]
\centering
\small
\begin{tabular}{lrrrrrrr}
\toprule
& POS & NER & XQuAD & MLQA & TyDiQA & XNLI & PAWS-X \\ \midrule
Batch size & \{8,16,32\} & 8 & 32 & 32 & 32 & 32 & 32 \\
Learning rate & \{1,2,3\}e-5 & \{5,...,9\}e-6 & \{2,3,4\}e-5 & \{2,3,4\}e-5 & \{2,3,4\}e-5 & \{5,...,8\}e-6 & \{8,9,10,20\}e-6 \\
LR schedule & Linear & Linear & Linear & Linear & Linear & Linear & Linear \\
Warmup & 10\% & 10\% & 10\% & 10\% & 10\% & 12,500 steps & 10\% \\
Weight decay & 0 & 0 & 0 & 0 & 0 & 0 & 0 \\
Epochs & 10 & 10 & 4 & \{2,3,4\} & \{10,20,40\} & 10 & 10\\
\bottomrule
\end{tabular}
\caption{Hyperparameters used for fine-tuning on the XTREME end tasks.}
\label{table:hparam}
\end{table*}

\end{document}

%% file: settings.tex
\usepackage{multirow}
\usepackage{amsmath}
\usepackage{capt-of}
\usepackage{tabularx}
\usepackage{epsfig}
\usepackage{amssymb}
\usepackage{amsfonts}
\usepackage{booktabs}
\usepackage{scalerel}
\usepackage[inline]{enumitem}
\usepackage{listings}
\usepackage{varwidth}
\usepackage[export]{adjustbox}
\usepackage{tikz}
\usetikzlibrary{tikzmark}
\usepackage{cleveref}

\usepackage{stmaryrd}
\usepackage{bbm}

\usepackage{algorithm}
\usepackage[noend]{algpseudocode}

\definecolor{deepblue}{rgb}{0,0,0.5}
\definecolor{officeblue}{RGB}{0,102,204}
\definecolor{deepred}{rgb}{0.6,0,0}
\definecolor{deepgreen}{rgb}{0,0.5,0}
\definecolor{mybrickred}{RGB}{182,50,28}

\definecolor{fillcolor}{RGB}{216,217,252}

\algnewcommand\algorithmicrequireb{{\hspace{0.85cm}}}
\algnewcommand\INPTDESCB{\item[\algorithmicrequireb]}

\algnewcommand\algorithmicfuncdesc{\textbf{Function:}}
\algnewcommand\FUNCDESC{\item[\algorithmicfuncdesc]}
\algnewcommand\algorithmicfuncdescb{{\hspace{1.48cm}}}
\algnewcommand\FUNCDESCB{\item[\algorithmicfuncdescb]}
\algnewcommand{\algorithmicgoto}{\textbf{goto}}
\algnewcommand{\Goto}[1]{\algorithmicgoto~\ref{#1}}



%% file: math_commands.tex
\usepackage{amsmath,amsfonts,bm}

\def\eqref#1{equation~\ref{#1}}

\def\1{\bm{1}}

\def\vtheta{{\bm{\theta}}}

\def\ve{{\bm{e}}}
\def\vf{{\bm{f}}}

\def\vx{{\bm{x}}}

\DeclareMathAlphabet{\mathsfit}{\encodingdefault}{\sfdefault}{m}{sl}
\SetMathAlphabet{\mathsfit}{bold}{\encodingdefault}{\sfdefault}{bx}{n}

\newcommand{\Ls}{\mathcal{L}}

\newcommand{\sigmoid}{\sigma}



%% file: main.bbl
\begin{thebibliography}{48}
\expandafter\ifx\csname natexlab\endcsname\relax\def\natexlab#1{#1}\fi

\bibitem[{Artetxe et~al.(2020)Artetxe, Ruder, and Yogatama}]{xquad}
Mikel Artetxe, Sebastian Ruder, and Dani Yogatama. 2020.
\newblock \href {https://doi.org/10.18653/v1/2020.acl-main.421} {On the
  cross-lingual transferability of monolingual representations}.
\newblock In \emph{Proceedings of the 58th Annual Meeting of the Association
  for Computational Linguistics}, pages 4623--4637, Online. Association for
  Computational Linguistics.

\bibitem[{Artetxe and Schwenk(2019)}]{tatoeba}
Mikel Artetxe and Holger Schwenk. 2019.
\newblock \href {https://transacl.org/index.php/tacl/article/view/1742}
  {Massively multilingual sentence embeddings for zero-shot cross-lingual
  transfer and beyond}.
\newblock \emph{Transactions of the Association for Computational Linguistics},
  7(0):597--610.

\bibitem[{Bao et~al.(2020)Bao, Dong, Wei, Wang, Yang, Liu, Wang, Gao, Piao,
  Zhou, and Hon}]{unilmv2}
Hangbo Bao, Li~Dong, Furu Wei, Wenhui Wang, Nan Yang, Xiaodong Liu, Yu~Wang,
  Jianfeng Gao, Songhao Piao, Ming Zhou, and Hsiao-Wuen Hon. 2020.
\newblock \href
  {https://proceedings.icml.cc/static/paper_files/icml/2020/3934-Paper.pdf}
  {{UniLMv2}: Pseudo-masked language models for unified language model
  pre-training}.
\newblock In \emph{Proceedings of the 37th International Conference on Machine
  Learning}, pages 7006--7016.

\bibitem[{Cao et~al.(2020)Cao, Kitaev, and Klein}]{Cao2020Multilingual}
Steven Cao, Nikita Kitaev, and Dan Klein. 2020.
\newblock \href {https://openreview.net/forum?id=r1xCMyBtPS} {Multilingual
  alignment of contextual word representations}.
\newblock In \emph{International Conference on Learning Representations}.

\bibitem[{Chi et~al.(2021{\natexlab{a}})Chi, Dong, Ma, Huang, Mao, Huang, and
  Wei}]{mt6}
Zewen Chi, Li~Dong, Shuming Ma, Shaohan Huang, Xian-Ling Mao, Heyan Huang, and
  Furu Wei. 2021{\natexlab{a}}.
\newblock {mT6}: Multilingual pretrained text-to-text transformer with
  translation pairs.
\newblock \emph{arXiv preprint arXiv:2104.08692}.

\bibitem[{Chi et~al.(2020)Chi, Dong, Wei, Wang, Mao, and Huang}]{xnlg}
Zewen Chi, Li~Dong, Furu Wei, Wenhui Wang, Xian{-}Ling Mao, and Heyan Huang.
  2020.
\newblock \href {https://www.aaai.org/Papers/AAAI/2020GB/AAAI-ChiZ.7682.pdf}
  {Cross-lingual natural language generation via pre-training}.
\newblock In \emph{The Thirty-Fourth {AAAI} Conference on Artificial
  Intelligence, {AAAI} 2020, New York, NY, USA, February 7-12, 2020}, pages
  7570--7577. {AAAI} Press.

\bibitem[{Chi et~al.(2021{\natexlab{b}})Chi, Dong, Wei, Yang, Singhal, Wang,
  Song, Mao, Huang, and Zhou}]{infoxlm}
Zewen Chi, Li~Dong, Furu Wei, Nan Yang, Saksham Singhal, Wenhui Wang, Xia Song,
  Xian-Ling Mao, Heyan Huang, and Ming Zhou. 2021{\natexlab{b}}.
\newblock \href {https://doi.org/10.18653/v1/2021.naacl-main.280} {{I}nfo{XLM}:
  An information-theoretic framework for cross-lingual language model
  pre-training}.
\newblock In \emph{Proceedings of the 2021 Conference of the North American
  Chapter of the Association for Computational Linguistics: Human Language
  Technologies}, pages 3576--3588, Online. Association for Computational
  Linguistics.

\bibitem[{Chi et~al.(2021{\natexlab{c}})Chi, Dong, Zheng, Huang, Mao, Huang,
  and Wei}]{xlmalign}
Zewen Chi, Li~Dong, Bo~Zheng, Shaohan Huang, Xian-Ling Mao, Heyan Huang, and
  Furu Wei. 2021{\natexlab{c}}.
\newblock \href {https://doi.org/10.18653/v1/2021.acl-long.265} {Improving
  pretrained cross-lingual language models via self-labeled word alignment}.
\newblock In \emph{Proceedings of the 59th Annual Meeting of the Association
  for Computational Linguistics and the 11th International Joint Conference on
  Natural Language Processing (Volume 1: Long Papers)}, pages 3418--3430,
  Online. Association for Computational Linguistics.

\bibitem[{Cho et~al.(2014)Cho, van Merri{\"e}nboer, Gulcehre, Bahdanau,
  Bougares, Schwenk, and Bengio}]{gru}
Kyunghyun Cho, Bart van Merri{\"e}nboer, Caglar Gulcehre, Dzmitry Bahdanau,
  Fethi Bougares, Holger Schwenk, and Yoshua Bengio. 2014.
\newblock \href {https://doi.org/10.3115/v1/D14-1179} {Learning phrase
  representations using {RNN} encoder{--}decoder for statistical machine
  translation}.
\newblock In \emph{Proceedings of the 2014 Conference on Empirical Methods in
  Natural Language Processing ({EMNLP})}, pages 1724--1734, Doha, Qatar.
  Association for Computational Linguistics.

\bibitem[{Clark et~al.(2020{\natexlab{a}})Clark, Choi, Collins, Garrette,
  Kwiatkowski, Nikolaev, and Palomaki}]{tydiqa}
Jonathan~H. Clark, Eunsol Choi, Michael Collins, Dan Garrette, Tom Kwiatkowski,
  Vitaly Nikolaev, and Jennimaria Palomaki. 2020{\natexlab{a}}.
\newblock \href {https://doi.org/10.1162/tacl_a_00317} {{T}y{D}i {QA}: A
  benchmark for information-seeking question answering in typologically diverse
  languages}.
\newblock \emph{Transactions of the Association for Computational Linguistics},
  8:454--470.

\bibitem[{Clark et~al.(2020{\natexlab{b}})Clark, Luong, Le, and
  Manning}]{electra}
Kevin Clark, Minh-Thang Luong, Quoc~V. Le, and Christopher~D. Manning.
  2020{\natexlab{b}}.
\newblock \href {https://openreview.net/forum?id=r1xMH1BtvB} {Electra:
  Pre-training text encoders as discriminators rather than generators}.
\newblock In \emph{International Conference on Learning Representations}.

\bibitem[{Conneau et~al.(2020)Conneau, Khandelwal, Goyal, Chaudhary, Wenzek,
  Guzm{\'a}n, Grave, Ott, Zettlemoyer, and Stoyanov}]{xlmr}
Alexis Conneau, Kartikay Khandelwal, Naman Goyal, Vishrav Chaudhary, Guillaume
  Wenzek, Francisco Guzm{\'a}n, Edouard Grave, Myle Ott, Luke Zettlemoyer, and
  Veselin Stoyanov. 2020.
\newblock \href {https://www.aclweb.org/anthology/2020.acl-main.747}
  {Unsupervised cross-lingual representation learning at scale}.
\newblock In \emph{Proceedings of the 58th Annual Meeting of the Association
  for Computational Linguistics}, pages 8440--8451, Online. Association for
  Computational Linguistics.

\bibitem[{Conneau and Lample(2019)}]{xlm}
Alexis Conneau and Guillaume Lample. 2019.
\newblock \href
  {http://papers.nips.cc/paper/8928-cross-lingual-language-model-pretraining.pdf}
  {Cross-lingual language model pretraining}.
\newblock In \emph{Advances in Neural Information Processing Systems}, pages
  7057--7067. Curran Associates, Inc.

\bibitem[{Conneau et~al.(2018)Conneau, Rinott, Lample, Williams, Bowman,
  Schwenk, and Stoyanov}]{xnli}
Alexis Conneau, Ruty Rinott, Guillaume Lample, Adina Williams, Samuel Bowman,
  Holger Schwenk, and Veselin Stoyanov. 2018.
\newblock \href {https://doi.org/10.18653/v1/D18-1269} {{XNLI}: Evaluating
  cross-lingual sentence representations}.
\newblock In \emph{Proceedings of the 2018 Conference on Empirical Methods in
  Natural Language Processing}, pages 2475--2485, Brussels, Belgium.
  Association for Computational Linguistics.

\bibitem[{Devlin et~al.(2019)Devlin, Chang, Lee, and Toutanova}]{bert}
Jacob Devlin, Ming-Wei Chang, Kenton Lee, and Kristina Toutanova. 2019.
\newblock \href {https://doi.org/10.18653/v1/N19-1423} {{BERT}: Pre-training of
  deep bidirectional transformers for language understanding}.
\newblock In \emph{Proceedings of the 2019 Conference of the North {A}merican
  Chapter of the Association for Computational Linguistics: Human Language
  Technologies, Volume 1 (Long and Short Papers)}, pages 4171--4186,
  Minneapolis, Minnesota. Association for Computational Linguistics.

\bibitem[{Dong et~al.(2019)Dong, Yang, Wang, Wei, Liu, Wang, Gao, Zhou, and
  Hon}]{unilm}
Li~Dong, Nan Yang, Wenhui Wang, Furu Wei, Xiaodong Liu, Yu~Wang, Jianfeng Gao,
  Ming Zhou, and Hsiao-Wuen Hon. 2019.
\newblock \href
  {http://papers.nips.cc/paper/9464-unified-language-model-pre-training-for-natural-language-understanding-and-generation.pdf}
  {Unified language model pre-training for natural language understanding and
  generation}.
\newblock In \emph{Advances in Neural Information Processing Systems}, pages
  13063--13075. Curran Associates, Inc.

\bibitem[{Dyer et~al.(2013)Dyer, Chahuneau, and Smith}]{fastalign}
Chris Dyer, Victor Chahuneau, and Noah~A Smith. 2013.
\newblock A simple, fast, and effective reparameterization of ibm model 2.
\newblock In \emph{Proceedings of the 2013 Conference of the North American
  Chapter of the Association for Computational Linguistics: Human Language
  Technologies}, pages 644--648.

\bibitem[{El-Kishky et~al.(2020)El-Kishky, Chaudhary, Guzm{\'a}n, and
  Koehn}]{el2019ccaligned}
Ahmed El-Kishky, Vishrav Chaudhary, Francisco Guzm{\'a}n, and Philipp Koehn.
  2020.
\newblock \href {https://doi.org/10.18653/v1/2020.emnlp-main.480} {{CCA}ligned:
  A massive collection of cross-lingual web-document pairs}.
\newblock In \emph{Proceedings of the 2020 Conference on Empirical Methods in
  Natural Language Processing (EMNLP)}, pages 5960--5969, Online. Association
  for Computational Linguistics.

\bibitem[{Fang et~al.(2022)Fang, Dong, Bao, Wang, and
  Wei}]{Fang2022CorruptedIM}
Yuxin Fang, Li~Dong, Hangbo Bao, Xinggang Wang, and Furu Wei. 2022.
\newblock Corrupted image modeling for self-supervised visual pre-training.
\newblock \emph{ArXiv}, abs/2202.03382.

\bibitem[{Goyal et~al.(2021)Goyal, Du, Ott, Anantharaman, and
  Conneau}]{xlmr-scaleup}
Naman Goyal, Jingfei Du, Myle Ott, Giri Anantharaman, and Alexis Conneau. 2021.
\newblock Larger-scale transformers for multilingual masked language modeling.
\newblock \emph{arXiv preprint arXiv:2105.00572}.

\bibitem[{Hao et~al.(2021)Hao, Dong, Bao, Xu, and Wei}]{hao-etal-2021-learning}
Yaru Hao, Li~Dong, Hangbo Bao, Ke~Xu, and Furu Wei. 2021.
\newblock \href {https://doi.org/10.18653/v1/2021.findings-acl.394} {Learning
  to sample replacements for {ELECTRA} pre-training}.
\newblock In \emph{Findings of the Association for Computational Linguistics:
  ACL-IJCNLP 2021}, pages 4495--4506, Online. Association for Computational
  Linguistics.

\bibitem[{Hu et~al.(2020{\natexlab{a}})Hu, Johnson, Firat, Siddhant, and
  Neubig}]{hu2020explicit}
Junjie Hu, Melvin Johnson, Orhan Firat, Aditya Siddhant, and Graham Neubig.
  2020{\natexlab{a}}.
\newblock Explicit alignment objectives for multilingual bidirectional
  encoders.
\newblock \emph{arXiv preprint arXiv:2010.07972}.

\bibitem[{Hu et~al.(2020{\natexlab{b}})Hu, Ruder, Siddhant, Neubig, Firat, and
  Johnson}]{xtreme}
Junjie Hu, Sebastian Ruder, Aditya Siddhant, Graham Neubig, Orhan Firat, and
  Melvin Johnson. 2020{\natexlab{b}}.
\newblock {XTREME}: A massively multilingual multi-task benchmark for
  evaluating cross-lingual generalization.
\newblock \emph{arXiv preprint arXiv:2003.11080}.

\bibitem[{Jalili~Sabet et~al.(2020)Jalili~Sabet, Dufter, Yvon, and
  Sch{\"u}tze}]{simalign}
Masoud Jalili~Sabet, Philipp Dufter, Fran{\c{c}}ois Yvon, and Hinrich
  Sch{\"u}tze. 2020.
\newblock \href {https://doi.org/10.18653/v1/2020.findings-emnlp.147}
  {{S}im{A}lign: High quality word alignments without parallel training data
  using static and contextualized embeddings}.
\newblock In \emph{Findings of the Association for Computational Linguistics:
  EMNLP 2020}, pages 1627--1643, Online. Association for Computational
  Linguistics.

\bibitem[{Joshi et~al.(2019)Joshi, Chen, Liu, Weld, Zettlemoyer, and
  Levy}]{spanbert}
Mandar Joshi, Danqi Chen, Yinhan Liu, Daniel~S Weld, Luke Zettlemoyer, and Omer
  Levy. 2019.
\newblock {SpanBERT}: Improving pre-training by representing and predicting
  spans.
\newblock \emph{arXiv preprint arXiv:1907.10529}.

\bibitem[{Kingma and Ba(2015)}]{adam}
Diederik~P. Kingma and Jimmy Ba. 2015.
\newblock \href {http://arxiv.org/abs/1412.6980} {Adam: {A} method for
  stochastic optimization}.
\newblock In \emph{3rd International Conference on Learning Representations},
  San Diego, CA.

\bibitem[{Kudo and Richardson(2018)}]{sentencepiece}
Taku Kudo and John Richardson. 2018.
\newblock \href {https://doi.org/10.18653/v1/D18-2012} {{S}entence{P}iece: A
  simple and language independent subword tokenizer and detokenizer for neural
  text processing}.
\newblock In \emph{Proceedings of the 2018 Conference on Empirical Methods in
  Natural Language Processing: System Demonstrations}, pages 66--71, Brussels,
  Belgium. Association for Computational Linguistics.

\bibitem[{Kunchukuttan et~al.(2018)Kunchukuttan, Mehta, and
  Bhattacharyya}]{iit}
Anoop Kunchukuttan, Pratik Mehta, and Pushpak Bhattacharyya. 2018.
\newblock \href {https://www.aclweb.org/anthology/L18-1548} {The {IIT} {B}ombay
  {E}nglish-{H}indi parallel corpus}.
\newblock In \emph{Proceedings of the Eleventh International Conference on
  Language Resources and Evaluation}, Miyazaki, Japan. European Language
  Resources Association.

\bibitem[{Lewis et~al.(2020)Lewis, Oguz, Rinott, Riedel, and Schwenk}]{mlqa}
Patrick Lewis, Barlas Oguz, Ruty Rinott, Sebastian Riedel, and Holger Schwenk.
  2020.
\newblock \href {https://www.aclweb.org/anthology/2020.acl-main.653} {{MLQA}:
  Evaluating cross-lingual extractive question answering}.
\newblock In \emph{Proceedings of the 58th Annual Meeting of the Association
  for Computational Linguistics}, pages 7315--7330, Online. Association for
  Computational Linguistics.

\bibitem[{Luo et~al.(2020)Luo, Wang, Liu, Liu, Bi, Huang, Huang, and Si}]{veco}
Fuli Luo, Wei Wang, Jiahao Liu, Yijia Liu, Bin Bi, Songfang Huang, Fei Huang,
  and Luo Si. 2020.
\newblock {VECO}: Variable encoder-decoder pre-training for cross-lingual
  understanding and generation.
\newblock \emph{arXiv preprint arXiv:2010.16046}.

\bibitem[{Ma et~al.(2021)Ma, Dong, Huang, Zhang, Muzio, Singhal, Awadalla,
  Song, and Wei}]{deltalm}
Shuming Ma, Li~Dong, Shaohan Huang, Dongdong Zhang, Alexandre Muzio, Saksham
  Singhal, Hany~Hassan Awadalla, Xia Song, and Furu Wei. 2021.
\newblock {DeltaLM}: Encoder-decoder pre-training for language generation and
  translation by augmenting pretrained multilingual encoders.
\newblock \emph{arXiv preprint arXiv:2106.13736}.

\bibitem[{Meng et~al.(2021)Meng, Xiong, Bajaj, Tiwary, Bennett, Han, and
  Song}]{cocolm}
Yu~Meng, Chenyan Xiong, Payal Bajaj, Saurabh Tiwary, Paul Bennett, Jiawei Han,
  and Xia Song. 2021.
\newblock {COCO-LM}: Correcting and contrasting text sequences for language
  model pretraining.
\newblock \emph{arXiv preprint arXiv:2102.08473}.

\bibitem[{Och and Ney(2003)}]{och2003systematic}
Franz~Josef Och and Hermann Ney. 2003.
\newblock A systematic comparison of various statistical alignment models.
\newblock \emph{Computational linguistics}, 29(1):19--51.

\bibitem[{Pan et~al.(2017)Pan, Zhang, May, Nothman, Knight, and Ji}]{panx}
Xiaoman Pan, Boliang Zhang, Jonathan May, Joel Nothman, Kevin Knight, and Heng
  Ji. 2017.
\newblock \href {https://doi.org/10.18653/v1/P17-1178} {Cross-lingual name
  tagging and linking for 282 languages}.
\newblock In \emph{Proceedings of the 55th Annual Meeting of the Association
  for Computational Linguistics (Volume 1: Long Papers)}, pages 1946--1958,
  Vancouver, Canada. Association for Computational Linguistics.

\bibitem[{Parikh et~al.(2016)Parikh, T{\"a}ckstr{\"o}m, Das, and
  Uszkoreit}]{parikh-etal-2016-decomposable}
Ankur Parikh, Oscar T{\"a}ckstr{\"o}m, Dipanjan Das, and Jakob Uszkoreit. 2016.
\newblock \href {https://doi.org/10.18653/v1/D16-1244} {A decomposable
  attention model for natural language inference}.
\newblock In \emph{Proceedings of the 2016 Conference on Empirical Methods in
  Natural Language Processing}, pages 2249--2255, Austin, Texas. Association
  for Computational Linguistics.

\bibitem[{Raffel et~al.(2020)Raffel, Shazeer, Roberts, Lee, Narang, Matena,
  Zhou, Li, and Liu}]{t5}
Colin Raffel, Noam Shazeer, Adam Roberts, Katherine Lee, Sharan Narang, Michael
  Matena, Yanqi Zhou, Wei Li, and Peter~J. Liu. 2020.
\newblock \href {http://jmlr.org/papers/v21/20-074.html} {Exploring the limits
  of transfer learning with a unified text-to-text transformer}.
\newblock \emph{Journal of Machine Learning Research}, 21(140):1--67.

\bibitem[{Rahimi et~al.(2019)Rahimi, Li, and Cohn}]{rahimi2019}
Afshin Rahimi, Yuan Li, and Trevor Cohn. 2019.
\newblock \href {https://doi.org/10.18653/v1/P19-1015} {Massively multilingual
  transfer for {NER}}.
\newblock In \emph{Proceedings of the 57th Annual Meeting of the Association
  for Computational Linguistics}, pages 151--164, Florence, Italy. Association
  for Computational Linguistics.

\bibitem[{Schwenk et~al.(2019)Schwenk, Chaudhary, Sun, Gong, and
  Guzm{\'a}n}]{wikimatrix}
Holger Schwenk, Vishrav Chaudhary, Shuo Sun, Hongyu Gong, and Francisco
  Guzm{\'a}n. 2019.
\newblock {WikiMatrix}: Mining {135M} parallel sentences in 1620 language pairs
  from {wikipedia}.
\newblock \emph{arXiv preprint arXiv:1907.05791}.

\bibitem[{Tiedemann(2012)}]{opus}
J{\"o}rg Tiedemann. 2012.
\newblock \href
  {http://www.lrec-conf.org/proceedings/lrec2012/pdf/463_Paper.pdf} {Parallel
  data, tools and interfaces in {OPUS}}.
\newblock In \emph{Proceedings of the Eighth International Conference on
  Language Resources and Evaluation}, pages 2214--2218, Istanbul, Turkey.
  European Language Resources Association.

\bibitem[{Vaswani et~al.(2017)Vaswani, Shazeer, Parmar, Uszkoreit, Jones,
  Gomez, Kaiser, and Polosukhin}]{transformer}
Ashish Vaswani, Noam Shazeer, Niki Parmar, Jakob Uszkoreit, Llion Jones,
  Aidan~N Gomez, {\L}ukasz Kaiser, and Illia Polosukhin. 2017.
\newblock \href
  {http://papers.nips.cc/paper/7181-attention-is-all-you-need.pdf} {Attention
  is all you need}.
\newblock In \emph{Advances in Neural Information Processing Systems}, pages
  5998--6008. Curran Associates, Inc.

\bibitem[{Xue et~al.(2021)Xue, Constant, Roberts, Kale, Al-Rfou, Siddhant,
  Barua, and Raffel}]{mt5}
Linting Xue, Noah Constant, Adam Roberts, Mihir Kale, Rami Al-Rfou, Aditya
  Siddhant, Aditya Barua, and Colin Raffel. 2021.
\newblock \href {https://doi.org/10.18653/v1/2021.naacl-main.41} {m{T}5: A
  massively multilingual pre-trained text-to-text transformer}.
\newblock In \emph{Proceedings of the 2021 Conference of the North American
  Chapter of the Association for Computational Linguistics: Human Language
  Technologies}, pages 483--498, Online. Association for Computational
  Linguistics.

\bibitem[{Yang et~al.(2020)Yang, Ma, Zhang, Wu, Li, and Zhou}]{alm}
Jian Yang, Shuming Ma, Dongdong Zhang, Shuangzhi Wu, Zhoujun Li, and Ming Zhou.
  2020.
\newblock Alternating language modeling for cross-lingual pre-training.
\newblock In \emph{Thirty-Fourth AAAI Conference on Artificial Intelligence}.

\bibitem[{Yang et~al.(2019{\natexlab{a}})Yang, Zhang, Tar, and
  Baldridge}]{pawsx}
Yinfei Yang, Yuan Zhang, Chris Tar, and Jason Baldridge. 2019{\natexlab{a}}.
\newblock \href {https://doi.org/10.18653/v1/D19-1382} {{PAWS}-{X}: A
  cross-lingual adversarial dataset for paraphrase identification}.
\newblock In \emph{Proceedings of the 2019 Conference on Empirical Methods in
  Natural Language Processing and the 9th International Joint Conference on
  Natural Language Processing (EMNLP-IJCNLP)}, pages 3687--3692, Hong Kong,
  China. Association for Computational Linguistics.

\bibitem[{Yang et~al.(2019{\natexlab{b}})Yang, Dai, Yang, Carbonell,
  Salakhutdinov, and Le}]{xlnet}
Zhilin Yang, Zihang Dai, Yiming Yang, Jaime Carbonell, Russ~R Salakhutdinov,
  and Quoc~V Le. 2019{\natexlab{b}}.
\newblock \href
  {https://proceedings.neurips.cc/paper/2019/file/dc6a7e655d7e5840e66733e9ee67cc69-Paper.pdf}
  {Xlnet: Generalized autoregressive pretraining for language understanding}.
\newblock In \emph{Advances in Neural Information Processing Systems},
  volume~32. Curran Associates, Inc.

\bibitem[{Zeman et~al.(2019)Zeman, Nivre, Abrams, and et~al.}]{udpos}
Daniel Zeman, Joakim Nivre, Mitchell Abrams, and et~al. 2019.
\newblock \href {http://hdl.handle.net/11234/1-3105} {Universal dependencies
  2.5}.
\newblock {LINDAT}/{CLARIAH}-{CZ} digital library at the Institute of Formal
  and Applied Linguistics ({{\'U}FAL}), Faculty of Mathematics and Physics,
  Charles University.

\bibitem[{Zhao et~al.(2021)Zhao, Eger, Bjerva, and
  Augenstein}]{zhao2020inducing}
Wei Zhao, Steffen Eger, Johannes Bjerva, and Isabelle Augenstein. 2021.
\newblock \href {https://doi.org/10.18653/v1/2021.starsem-1.22} {Inducing
  language-agnostic multilingual representations}.
\newblock In \emph{Proceedings of *SEM 2021: The Tenth Joint Conference on
  Lexical and Computational Semantics}, pages 229--240, Online. Association for
  Computational Linguistics.

\bibitem[{Zheng et~al.(2021)Zheng, Dong, Huang, Singhal, Che, Liu, Song, and
  Wei}]{vocap}
Bo~Zheng, Li~Dong, Shaohan Huang, Saksham Singhal, Wanxiang Che, Ting Liu, Xia
  Song, and Furu Wei. 2021.
\newblock \href {https://aclanthology.org/2021.emnlp-main.257} {Allocating
  large vocabulary capacity for cross-lingual language model pre-training}.
\newblock In \emph{Proceedings of the 2021 Conference on Empirical Methods in
  Natural Language Processing}, pages 3203--3215, Online and Punta Cana,
  Dominican Republic. Association for Computational Linguistics.

\bibitem[{Ziemski et~al.(2016)Ziemski, Junczys-Dowmunt, and
  Pouliquen}]{multiun}
Micha{\l} Ziemski, Marcin Junczys-Dowmunt, and Bruno Pouliquen. 2016.
\newblock The united nations parallel corpus v1. 0.
\newblock In \emph{LREC}, pages 3530--3534.

\end{thebibliography}
